# Deep Neural Networks for the Assessment of Surgical Skills: A Systematic Review

**Erim Yanik[1], Xavier Intes[2], Uwe Kruger[2], Pingkun Yan[2], David Diller[3], Brian Van Voorst[3], Basiel Makled[4], Jack Norfleet[4], Suvranu De[1]**

[1]Department of Mechanical, Aerospace, and Nuclear Engineering, Rensselaer Polytechnic Institute, USA

[2] Department of Biomedical Engineering, Rensselaer Polytechnic Institute, USA

[3]Raytheon BBN Technologies, USA

[4]Simulation and Training Technology Center, Army Research Laboratory, USA

**One Sentence Summary**

This paper provides a comprehensive review of the current state of the literature on automated surgical skill assessment based on deep neural networks (DNN).

**Abstract**

Surgical training in medical school residency programs has followed the apprenticeship model. The learning and assessment process is inherently subjective and time-consuming. Thus, there is a need for objective methods to assess surgical skills. Here, we use the Preferred Reporting Items for Systematic Reviews and Meta-Analyses (PRISMA) guidelines to systematically survey the literature on the use of Deep Neural Networks for automated and objective surgical skill assessment, with a focus on kinematic data as putative markers of surgical competency. There is considerable recent interest in deep neural networks (DNN) due to the availability of powerful algorithms, multiple datasets, some of which are publicly available, as well as efficient computational hardware to train and host them. We have reviewed 530 papers, of which we selected 25 for this systematic review. Based on this review, we concluded that DNNs are powerful tools for automated, objective surgical skill assessment using both kinematic and video data. The field would benefit from large, publicly available, annotated datasets that are representative of the surgical trainee and expert demographics and multimodal data beyond kinematics and videos.

## 1. Introduction

Surgical procedures require high dexterity, fine motor control, and hand-eye coordination. Surgical training has followed the Halstedian apprenticeship model of learning with an expert in the operating room (OR).[1,2] Traditional surgical assessment methods, such as direct observations by an experienced trainer to assess the skills of the trainee, are generally subjective and use global rating scales (GRS), e.g., the Objective Structure Assessment of Technical Skills (OSATS), to score competency.[3] These observational methods allow experienced surgeons to use structured checklists for technical criteria and rate the surgical performance of the trainee under direct observation. Surgical training programs also utilize other checklist-based approaches, e.g., In-Training Evaluation Reports (ITERs), to evaluate trainees. These approaches suffer from the well-known limitations of being subjective with poor inter-rater reliability, distribution errors, recall bias, and halo effects.[4-6]

Increasingly, simulators are being used to provide an efficient training environment outside the OR and certify surgeons, eliminating risks to human patients. Notably, the Fundamentals of Laparoscopic Surgery (FLS) program, developed by a joint committee of the American Gastrointestinal and Endoscopic Surgeons (SAGES) and the American College of Surgeons (ACS), is a pre-requisite for board certification in general and ob/GYN surgery.[7,8] The Fundamentals of Endoscopic Surgery (FES) utilizes a virtual reality-based simulator and is required for general surgery board certification.[9] Moreover, a more recent curriculum, namely, Fundamentals of Robotic Surgery (FRS) offers standardization to assess surgeons in robotic surgery.[10] Simulators such as these use metrics based on time and error to provide performance scores to certify surgeons. The definition of error in these simulators is subjective and may not correlate with operative errors. The evaluation of these errors, e.g., calculating the deviation of a cut-trajectory from a pre-marked circle in the pattern cutting task of the FLS is also manually intensive and subjective. Also, time to completion is heavily weighted, with no evidence that completing a task faster is indicative of surgical proficiency.

There is currently an intense focus on developing data-driven techniques to overcome the limitations of surgical skill assessment. Two of the major categories of data that are collected for skills assessment include kinematic and video-based data. Kinematic data is collected by placing sensors at tooltips, hand, or on other locations of the body of the trainee to collect motion-related data.[12-16] Data for features such as location, velocity, and rotation of a surgical tool can be collected directly using these sensors. This type of data collection is invasive due to the presence of the sensor that might interfere with the motion of the trainees. The kinematic data can also be collected virtually using software such as EndoVis for virtual training applications, which does not suffer from the limitations of physical sensors.[17] Unlike kinematic data, video-based data is gathered in a non-invasive manner and is easier to work with. However, video-based data requires extensive post-processing to extract meaningful information. Crowdsourcing has been proposed as a method to leverage the wisdom of the crowd and reduce the need for time-intensive work by medical professionals for scoring surgical videos.[18]

Automated data processing is a pre-requisite to automating skill assessment. Earlier works derive statistical representations from the collected data to represent surgical performance.[19-21] Anh *et*

*al.*[22] used the mean, min-max-standard deviation (SD) of each sequence together with more complex representations, e.g., the average absolute difference, the average root sum of squared level, the average root square difference, and binned distribution to represent the kinematic tool motion data collected for elementary surgical tasks performed using surgical robots.[13] With the statistical features of choice, they were able to differentiate experts from novices with accuracies between 87.25% and 95.42%. Moreover, Kim *et al.*[23] employed crowd-sourced tool tip velocities (TTV), and optical flow fields (OFF) to represent tool motion during capsulorhexis and were able to differentiate between experts and novices with accuracies of 84.8% and 63.4% for TTV and OFF, respectively.

Although statistical representations prove useful in specific scenarios, manual extraction of the features, needed for domain knowledge, and time-effectiveness are critical limitations. Machine learning (ML) techniques, including hidden Markov models (HMM), support vector machines (SVM), and bag of words (BoW) are data-driven models, extensively used for surgical skill assessment.[25-28] In a hybrid study, Fard *et al.*[29] extracted statistical features, including duration, path length, and smoothness, from the kinematic tool motion data provided by the DaVinci surgical robot (dVSS). The features were selected to represent the surgeon's dexterity based on tool motion. Logistic regression (LR) and support vector machine (SVM) were applied for binary classification of eight surgeons performing suturing, and a global rating score was utilized to annotate the surgeon's performance. As a result, a maximum of 85.7% overall accuracy was reported using LR with leave-one-super-trial-out (LOSO) cross-validation (CV). In another study conducted by Zia *et al.*[31], various ML techniques were compared, including HMM, BoW, Motion Textures, and Discrete Cosine Transform (DCT), which performs well with repetitive tasks. They collected video-based data on suturing and knot-tying and used OSATS to establish ground-truth for the performance. As a result, they showed the benefit of using frequency domain-based techniques, i.e., DCT, over the techniques mentioned above for repetitive tasks.

While ML techniques are promising, they rely on manual features that are task-specific and require significant effort to define and optimize. Also, neither statistical features nor ML techniques are end-to-end, i.e., they need multiple steps to produce the desired output, e.g., extracting information from the collected data, computing features, and then using the ML model or an equation to assess performance. Deep learning (DL), is rapidly emerging as a method of choice to overcome these limitations of ML approaches.[32] DL is a powerful supervised learning technique that consists of hidden layers with neurons within. The neurons extract and represent complex features, and the complexity can be enhanced by adding more layers and neurons. The utility of DL is used to identify surgical tools in every frame of surgical videos, automatically, based on the extractedfeatures that are otherwise non-detectible to human eye.

The availability of powerful algorithms, large labeled datasets, and inexpensive high-power computing is fueling the growth in DL-based approaches in surgical assessment[33,34]. In recent years complex recurrent neural networks (RNNs) such as Long Short Term Memory (LSTM), Gated Recurring Unit (GRU), and their combinations with convolutional neural networks (CNN) have been utilized for surgical skill assessment.[35,36] CNN's were also used to assess skills and generate meaningful feedback to the trainee.[37-40] Moreover, CNNs such as Hourglass Network,

Faster Region-Based CNN (Faster RCNN), and Temporal Segment Networks (TSN), and others were used to extract tool locations from surgical videos.[41-45]

In this paper, we systematically analyzed automated and objective surgical skill assessment techniques based on deep neural networks (DNN) using the Preferred Reporting Items for Systematic Reviews and Meta-Analyses (PRISMA) guidelines.[46] While there are multiple review papers that have been recently published in similar areas, they lack technical analysis on DNNs and do not state the benefits of DNNs over traditional machine learning techniques for surgical skill assessment, such as visual feedback or end-to-end learning.[33,34]

The rest of the paper is organized as follows. The review criteria are presented in Section 2. Section 3 discusses the major findings from the review, including the available datasets, deep learning models utilized, prevalent cross-validation methods, and skill assessment techniques. Sections 4 and 5 present discussions and conclusions, respectively.

## 2. Review criteria

We used Google Scholar as the database and included all the published works through June 2020 in the analysis. We did not have a designated start date and the earliest work we could find in this field belonged November 2013. Eight different keywords and fourteen different keyword combinations were used to identify the 527 potential publications. The keywords related to surgical skills, namely "surgical skill assessment," "surgical skill classification," and "surgical skill evaluation," were used together with deep learning terminology, including "deep learning," "artificial intelligence," "deep neural networks," LSTM," and "convolutional neural networks". Each keyword combination was treated individually. Papers cited in the obtained papers were also included in our list. We excluded 364 papers that were duplicates, and patents, books, theses, and publications in foreign languages. The screening was done by reading the title and the abstract of each paper. The papers that were excluded at this stage were mostly related to different applications of surgery or were not utilizing DL techniques. We believe that these papers were included in the repository due to the articles they cited. After the screening process, we identified 46 papers for an extensive review. In this step, we read each paper thoroughly. We removed the papers about tool detection & localization, workflow and action recognition, and video segmentation in surgery using DNNs. The reason is that even though these steps are required for skill assessment, the authors did not utilize their technology to do so. Furthermore, some studies that were eliminated reported results by using ML techniques and claimed that DL is their next study aim. This process finally led to a total of 25 publications fulfilling all criteria (Figure 1). The temporal distribution of the papers is presented in Figure 2.

## 3. Review results

We observed that the published papers on DNN-based surgical skill assessment could be broadly categorized based on the dataset used – those that utilized the JHU-ISI Gesture and Skill

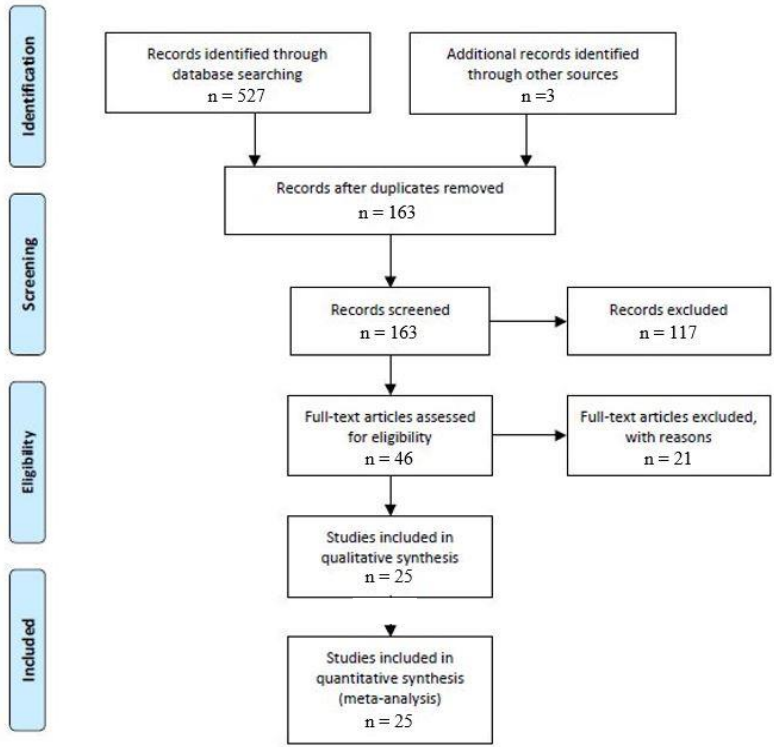

**Figure 1.** PRISMA flowchart for study selection.

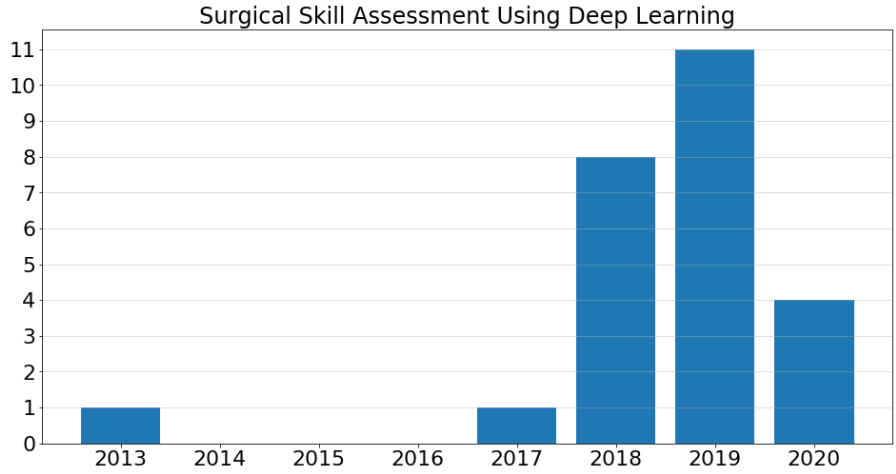

**Figure 2.** Published papers that were chosen for this review.

Assessment Working Set (JIGSAWS) dataset and those that used other datasets.[13] The papers also differed in the data acquisition, DNN models, cross-validation, and skill assessment techniques, as presented below and summarized in Tables 1 and 2.

**Table 1.** Summary of objective and automated surgical skill assessment on JIGSAWS dataset

| Author [Year] | Data Acquisition | Data Processing | CV | Form. | Visual-Feedback |
|---|---|---|---|---|---|
| Benmansour et al.[47] [2018] | Kinematic | DNN | N/A | C | N/A |
| Doughty et al.[48] [2018] | Video-Based | Sesame Model + Two stream TSN | 4-fold | C | N/A |
| Fawaz et al.[40] [2018] | Kinematic | CNN with GAP | LOSO | C | CAM |
| Wang & Fey[37] [2018] | Kinematic | CNN | LOSO & Holdout | C | N/A |
| Wang & Fey[49] [2018] | Kinematic | CNN-GRU | LOSO | C | N/A |
| Anh et al.[22] [2019] | Kinematic | **CNN, CNN-LSTM, LSTM, PCA, Autoencoder. Statistical. DFT. DCT** | LOSO | C | N/A |
| Benmansour et al.[50] [2019] | Kinematic | 3 RNNs for KT, NP, and ST | N/A | C | N/A |
| Castro et al.[39] [2019] | Kinematic | CNN with SELU + GMP + Quaternion Convolutions | LOSO | C | CAM |
| Fawaz et al.[38] [2019] | Kinematic | CNN with GAP | LOSO | R&C | CAM |
| Funke et al.[43] [2019] | Video-Based | 3DCNN + TSN (First end-to-end) | LOSO & LOUO | C | N/A |
| Li et al.[51] [2019] | Video-Based | RNN-based attention network | 4-fold | C | AM |
| Nguyen et al.[52] [2019] | Kinematic | CNN-LSTM | LOSO | C | N/A |
| Zhang et al.[53] [2020] | Kinematic | CNN | LOSO | C | CAM |

GAP: Global Average Pooling, GMP: Global Mixed Pooling, CAM: Class Activation Function, AM: Attention Maps,
KT: Knot Tying, NP: Needle Passing, ST: Suturing
CV: Cross-Validation, C: Classification, R: Regression
**Note**: Bold-written models under Data Processing achieve the highest score

**Table 2a.** Summary of objective and automated surgical skill assessment on *kinematic* data

| Author [Year] | Surgical Dataset / Procedure | Type of Surgery | Data Processing | CV | Form. |
|---|---|---|---|---|---|
| Ahmad et al.[54] [2013] | PT & Grasping | Virtual | Elman DNN | N/A | R |
| Cifuentes et al.[55] [2019] | Transvaginal assessment diagnosis | Laparoscopic | **LSTM**, Simple RNN | N/A | C |
| Hung et al.[56] [2019] | RARP | Robotic | DeepSurv, DNN | 5-fold | C |
| Kowalewski et al.[15] [2019] | ST & KT | Laparoscopic | DNN | 10-fold | R & C |
| Pérez-Escamirosa et al.[57] [2019] | PC, ST, & KT | Virtual | RBFNet (feed forward DNN) | LOUO, Holdout | C |
| Uemura et al.[58] [2019] | ST | Laparoscopic | 9 statistical features + Chaos Neural Network | N/A | C |
| Getty et al.[59] [2020] | P&P, PT, & TOR | Virtual | **LSTM**, CNN, FCN (converted SNN) | N/A | C |
| Vaughan & Gabrys[60] [2020] | Epidural Training | Virtual | **ResNet**, FCN (**with GAP**), CNN with final discriminative layer, Multi-channel CNN | LOOCV, 5-fold | C |

SNN: Spiking Neural Network, RBFNet: Radial basis function networks
RARP: Robot-assisted radical prostatectomy, PT: Peg Transfer, ST: Suturing, KT: Knot Tying, PC: Pattern Cutting,
P&P: Pick and Place, TOR: Thread the Ring**s**
R: Regression, C: Classification
**Note**: Bold-written models under Data Processing achieve the highest score.

**Table 2b.** Summary of objective and automated surgical skill assessment on *video-based* data

| Author [Year] | Surgical Dataset / Procedure | Type of Surgery | Data Acquisition | Data Processing | CV | Form. |
|---|---|---|---|---|---|---|
| Law et al.[41] [2017] | Vesicourethral Anastomosis | Robotic | Hourglass network (CNN) | SVM | LOOCV | C |
| Jin et al.[42] [2018] | m2cai16-tool + m2cai16-tooL-locations | Laparoscopic | Faster RCNN | SFR | Splitting | C |
| Mazomenos et al.[61] [2018] | Transesophageal Echocardiography | Virtual | AlexNet & **VGG** | SFR | N/A | R & C |
| Kim et al.[23] [2019] | Capsulorhexis | Regular | TCN | SFR | 5-fold | C |

TCN: Temporal Convolutional Neural Network, RCNN: Region-Based Convolutional Neural Network
SFR: Statistical Feature Representation
R: Regression, C: Classification
**Note**: Bold-written models under Data Processing achieve the highest score

## 3.1. Datasets

The availability of relatively large datasets, including publicly available ones, has spurred significant growth of DNL-based techniques. Both kinematic and video-based data were collected and utilized in these datasets. The kinematic data was gathered using sensors, simulators, and/or surgical robots, whereas the video-based data were obtained either using virtual simulators or cameras. For instance, JIGSAWS is a publicly available dataset that is used in multiple studies.[13] It is a robotic surgery dataset containing data from three elementary tasks performed using the da Vinci surgical system (dVSS), including suturing, needle passing, and knot-tying. Suturing is done by passing a needle through a synthetic tissue. Needle passing, like suturing, is where a needle is passed through metal hoops from one side to the other. Knot-tying is when the subject ties a single loop knot using the ends of the suture. There are 39 suturing, 36 knot-tying, and 28 needle passing trials conducted by 8 subjects, which includes two experts with 100+ hours of experience, four novices with less than 10 hours of experience, and two intermediates with between 10 and 100 hours of experience. Here, the hours of expertise are self-proclaimed and expert and intermediate subjects are underrepresented in comparison to novices. In addition, each trial is annotated using a modified version of OSATS. The dataset includes a total of 76 variables representing kinematic data associated with tooltip positions in Cartesian coordinates, tooltip rotation matrix, tip linear and rotational velocities, and gripper angle velocity for four robotic arms, including two master and two patient side manipulators. The kinematic data has been collected by dVSS along with video-based data recorded by endoscopic cameras at 30 FPS at a resolution of 640 x 480. The dataset has gesture-based verbal annotations yet does not possess spatial annotation for the tools used during the tasks.

Another publicly available dataset is Cholec80, which consists of 80 videos of laparoscopic cholecystectomy performed by 13 experts and captured at 25 FPS.[44] The binary tool presence annotation was provided at 1 FPS rate, offering a suitable environment to study DNN for tool detection.[62] In addition, mc2cai16-tool and mc2ai16-workflow are other datasets constructed based on Cholec80 for the 2016 M2CAI Challenge.[44,63] Mc2cai16-tool was later expanded by Jin et al.[42,] who introduced spatial annotation for the tools so that the dataset would be suitable for tracking tool motion and the technical assessment of surgical skills. They were also the first ones to use the dataset. Their results matched with OSATS classification for four experts. Other than public datasets, studies reported assessment on private datasets collected using the dVSS, simulators, sensors, and cameras.[54-56,58,60]

Following our analysis, we found 5 studies out of 25 that acquired data virtually using simulators. In addition, 15 papers, including the JIGSAWS studies, were robotic, and four were of laparoscopic surgery. One paper studied Capsulorhexis hence, labeled as general surgery. Based on works besides JIGSAWS, we found that half of the papers utilizing kinematic data acquire it virtually. In contrast, only a quarter of the papers obtained data from virtual simulations for video-based studies. Additionally, our survey showed that only tool motion (23 studies) and hand motion (2 studies) were utilized. We could not find papers employing other methods, including eye motion tracking and muscle contractions used in earlier ML techniques.[64] Between tool and hand motion, tool motion was favored as 92% of the papers are based on tracking tool motion to assess skills.

## 3.2. DNN models and data processing

There are three main DNN types used for surgical skill assessment. These are CNNs, RNNs, and Fully-Connected Neural Networks (FCN). Figure 3 illustrates the DNN models that have been utilized in the reviewed papers:

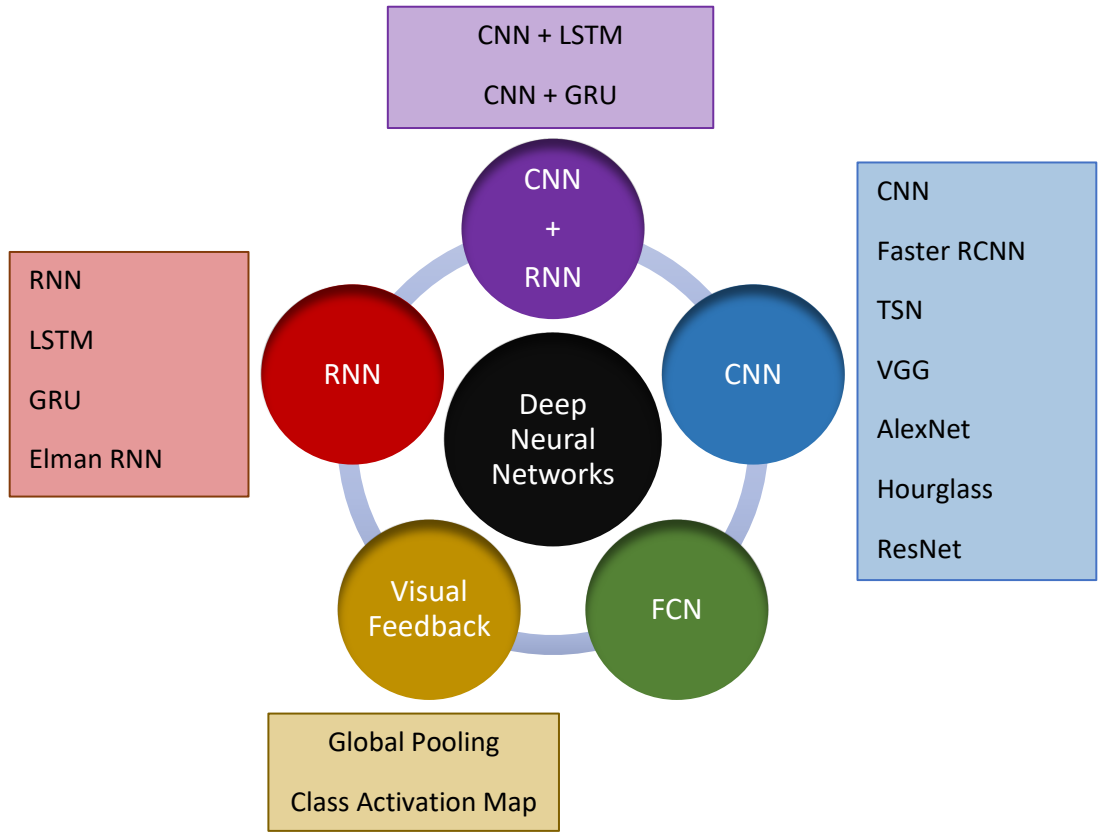

**Figure 3:** DNNs employed for surgical skill assessment. Circles other than the yellow represent the architectures, namely convolutional neural networks (CNN), recurrent neural networks (RNN), fully-connected networks (FCN), and CNN-RNN variates, whereas the yellow circle is for visualization. The color-coded boxes illustrate the specific frameworks used.

In the figure, the aforementioned DNNs and also combinations of CNNs and RNNs are shown. In addition, well-known employed structures such as VGG, ResNet, and LSTM are also mentioned. Class Activation Maps (CAM) are utilized to provide visual feedback to the trainee by highlighting the parts of the performance that contributed significantly to the outcome. They are mentioned under Visual Feedback in Figure 3. CAM is enabled through Global Pooling. Thus, it is also placed under Visual Feedback. Besides allowing visualization, Global Pooling also eliminates the duration difference problem that is crucial for time-series data.[38,39]

Two main data processing approaches were observed. For the kinematic inputs, the data can directly be fed to the DNN, and the model outputs either the regressed score or the computed class. As for the video-based inputs, data processing is usually a two-step process. A DL model first localizes either the tool or the hand motion, which is subsequently utilized by a second DNN to assess skill. Notwithstanding, there is one study where the authors developed a Temporal Segment Networks (TSN) as a Siamese architecture.[43,65] Multiple CNNs working on different segments of a video were aggregated in the consensus layer for classification. To the best of our knowledge, this is the first and only end-to-end model for video-based surgical skill assessment using DNN.

Additionally, out of the 25 papers, there were 12 CNN, 6 RNN, 3 CNN-RNN combinations, 3 Temporal-CNN, and 5 FCN models. Hence, CNN and CNN-X models are the most popular choice, being used in 62% of the studies. Some papers had more than one network developed for comparison. We also observed that the highest accuracies were obtained using CNN models. The results for the JIGSAWS dataset are given in Table 3. In this table, the first two rows are studies utilizing ML models, provided for comparison. The results by Zia *et al.* are the current state-of-the-art (Acc. 99.96%) for both ML and DNN, with the best overall accuracy. When the results are compared exclusively for DNN, the model introduced by Castro *et al.*[39] has the highest accuracy overall (98.7%) and also for Knot Tying (98.9%) whereas, for Suturing Funke *et al.*[43] (100%) and Fawaz *et al.*[38,40] (100%) and for Needle Passing Fawaz *et al.*[38,40] (100%) report the highest accuracies.

**Table 3.** Comparison of DNN accuracies based on JIGSAWS dataset / Study 1. (LOSO-CV)

| Author | Method | Suturing | Needle Passing | Knot Tying | Overall |
|---|---|---|---|---|---|
| Tao *et al.*[66] | S-HMM | 97.4 | 96.2 | 94.4 | 96.0 |
| Zia *et al.*[27] | ApEn | **100** | **100** | **99.9** | **99.96** |
| Zhang *et al.*[53] | CNN | 98.4 | 86.8 | 80.8 | 88.7 |
| Anh *et al.*[22] | LSTM | 95.1 | 91.5 | 89.6 | 92.1 |
| Wang *et al.*[37] | CNN | 92.5 | 95.4 | 91.3 | 93.1 |
| Anh *et al.*[22] | CNN-LSTM | 96.4 | 93.4 | 91.0 | 93.6 |
| Anh *et al.*[22] | CNN | 96.8 | 95.4 | 92.7 | 95.0 |
| Wang *et al.*[49] | CNN-GRU | N/A | N/A | N`/A | 96.6 |
| Nguyen *et al.*[52] | CNN-LSTM | 98.4 | 98.4 | 94.8 | 97.2 |
| Fawaz *et al.*[38,40] | CNN | **100** | **100** | 92.1 | 97.4 |
| Funke *et al.*[43] | 3DCNN+TSN | **100** | 96.4 | 95.8 | 97.4 |
| Castro *et al.*[39] | CNN | 98.4 | 98.9 | **98.9** | **98.7** |

## 3.3. Cross-validation

In ML and DL, cross-validation (CV) is a well-known technique to train a network using as many observations as possible and simultaneously validate the model performance.[30,67,68] The validation set does not contain observations that were used for model training. This is achieved by segmenting the data sets into k-folds and applying a holdout strategy, where one of the segments is removed. The network is trained using the remaining (k–1) folds, and the model performance on the fold left out is then recorded. The CV procedure methodically removes one fold at the time until each fold has served as a validation set. The 25 articles surveyed here apply three distinct CV strategies, which are as follows:

- **Leave-one-super-trial-out** (LOSO): The $i^{th}$ trial from each participant is removed and used to validate the performance of the DNN model while the remaining data is used for training. The performance of the DNN model is then evaluated after k-folds of super trials have been left out for model validation. k is 5 for JIGSAWS dataset.[13]
- **Leave-one-user-out** (LOUO): removing the trials conducted by a single user to validate the model performance until each user has been left out once.
- **k-fold cross-validation**, here the trials are concatenated in a randomized fashion and the combined dataset is divided into k folds. The aim of a randomized arrangement of trials is to avoid biasing the model validation to particular users.

LOSO is employed for studies utilizing the JIGSAWS dataset and those involving robotic surgery. 9 out of 25 papers have utilized LOSO to cross-validate their models. It is because JIGSAWS lacks sufficient data, and LOSO allows DNN to observe all but one data point for each participant during training. Moreover, LOSO achieved higher accuracy for skewed datasets such as, JIGSAWS. LOUO, which is another CV technique, have only been used to compare it with the other CV schemes. In addition, 7 out of 25 papers used k-folds (k=4, 5, & 10) to cross-validate their models. Here, two out of 7 was LOOCV, which is a special version of k-fold where k is equal to the data size. Furthermore, k-fold is the most popular CV method simultaneously amongst non-JIGSAWS and video-based skill assessment studies. Other than those above, four studies utilized Train/Test/Validation Set Splits, and one paper did not provide a validation during the training. Also, 6 out of 25 did not report any CV method. The models that were not cross-validated were trained and tested once; hence, there is no guarantee that the fixed set covers all possible variations of movements.

### 3.4. Skills assessment techniques

We found that the two skills assessment techniques used in this field are classification and regression. Out of the 25 papers analyzed, we found three papers that use regression in addition to classification, and only one used regression alone. For instance, in a study based on the JIGSAWS kinematic data, Fawaz et al.[38,40] used a fully connected neural network (FCN) to predict the OSATS score for each trial. They used Spearman's correlation coefficient to assess the model performance and computed 0.6, 0,57, and 0.65 for Suturing, Needle Passing, and Knot Tying, respectively. Secondly, Kowalewski et al.[15] also developed an FCN to predict OSATS score for Suturing and Knot Tying and reported an overall mean squared error (MSE) of 3.71 ± 0.64 and $R^2$ of 0.03 ± 0.81. They also did binary classification analysis and computed 70% accuracy.

For the classification studies, the papers based on the JIGSAWS datasets favored classification into three categories - Novice, Intermediate, and Expert. The Intermediate class suffers the most from misclassification. For example, Funke et al.[43] misclassified all the intermediate surgeons into either expert or novice using optical flow data with 3D-CNN for knot tying. Their model was cross-validated by LOUO, and they claimed that the uneven distribution of the dataset leads the DNN to fail to generalize the features. The lack of a sufficient number of expert surgeons might also explain why we did not find any study using LOUO alone. In another study, Wang and Fey[37] computed their lowermost Intermediate class score as 0.57 for knot-tying. Furthermore, the Intermediate class score was consistently lower for the other two JIGSAWS tasks. Anh *et al.*[22] also had similar performance issues for the Intermediate class, especially for Knot Tying.

In addition to these studies, Getty *et al.*[59] took another approach and classified four expert surgeons with similar expertise. They achieved 83.4% accuracy using LSTM. Also, three studies trained DNN only with expert data and ranked the non-expert surgeons based on the deviation.[47,50,54] The most common metrics to formulate regression were $R^2$ and Root Mean Squared Error (RMSE) whereas, it was accuracy, precision, recall, and F1 score for classification.

## 4. Discussion

Automation of technical skill assessment has the promise of objective evaluation, reduction of resource requirements, and personalized training with real-time feedback. In this study, we analyzed the current literature on the use of DNN for objective surgical skill assessment. Our first finding was that only tool and hand motion studies were available in the literature that fit our criteria. This may be attributed to the lack of public datasets or resources available for other data.

We also observed that kinematic and video-based data are the only two types of data utilized in the studies, with kinematic data being used in two-thirds of the published papers. We believe this is due to the fact that the sensors can record the kinematic data of the tool and hand motion directly, as opposed to video-based data. Moreover, it can be directly be fed to the DNN without any intermediate steps. Video-based data, on the other hand, requires an additional preprocessing step to locate the tools and hands. Authors either annotated certain portion of the dataset spatially or relied on crowdsourcing perform the same task. Amazon Mechanical Turk was the method of crowdsourcing for all the surveyed studies that utilized crowdsourcing. Then, DNNs were trained respectively, to detect and localize the tools for the rest of the dataset. However, collecting kinematic data is easier in robotic surgery or in virtual reality-based simulators and not in open or laparoscopic surgery, where tools will have to be instrumented during the operation. In addition, the features extracted from the kinematic data, such as tooltip location or rotation, might not cover the information that the collected data possess; hence, might not necessarily be optimal for the given surgical task. On the other hand, video data is easy to collect and unobtrusive and collected during laparoscopic and robotic surgeries. Although video-based data requires significant annotation and preprocessing and the localization DNNs require high computational power, they can be used to assess skills solemnly based on the recordings of performance without any explicit preparation once the network is trained and cross-validated.

Among the different techniques employed, CNNs were found to be superior to the other DNNs such as LSTM and FCN. This might be because RNNs are optimized for data sequences, and the public datasets are not rich with long sequences of surgical data. In addition, most CNNs take statistical features as input, which eliminates the need for LSTM since the temporal information is represented by the features of choice. Among the studies sharing the same dataset, JIGSAWS, we found that a model proposed by Castro et al.[39] achieved the highest overall score. They achieved 98.7% accuracy using a CNN with SELU activation, GMP, and quaternion convolutional layers. However, the current state-of-the-art for JIGSAWS dataset is due to Zia et al.[27] in which the predictability of the tool motion is encoded using approximate entropy [ApEn].[24] They reported 99.96% overall accuracy in differentiating surgeons into categories. Even though there is 1.26% difference in accuracy between these two models, we believe we should also consider two main advantages that Castro et al.[39] have over Zia et al.[27]. First, the ML model needs to extract information from the kinematics before feeding them into the nearest neighbor classifier. But, in Castro et al.[39], the kinematics are fed directly into CNN without an explicit step; therefore, it is an end-to-end model. Second, Castro et al.[39] embed the CAM to the DNN based on the work of Fawaz et al.[38,40] It fits the time-series data into a color segregated diagram with respect to the contribution of each timestamp. This visual feedback helps the participant on which parts of the performance he/she should improve. This is a quality missing in Zia et al.[27]. We believe these advantages are more than enough to favor the work of Castro et al.[39] over Zia et al.[27] especially considering the increasing amount of studies published over the years, as seen in Figure 3.

Our analysis suggested that LOSO is the most commonly used CV method for surgical skill assessment. Results obtained by LOSO are reported to be higher in comparison to LOUO. We believe this is because of the skewed nature of the datasets. For instance, JIGSAWS only has two experts and two intermediates, whereas there are four novices present. In LOSO the model is trained by seeing data from all the surgeons thus is exposed to all different styles. But, the same is not valid for LOUO because the DNN does not see data from one surgeon, and in a case the surgeon is an expert or an intermediate, the DNN is left with only one expert or intermediate data to train with. This has an adversarial effect on the generalization power of the DNN. In addition, models cross-validated by LOSO have already achieved accuracies up to 98.7%. Therefore, the field for LOSO CV is almost saturated. We hypothesized that, even though LOUO reports poorer scores, we will see more papers published to improve LOUO to compensate for the gap in the literature. We were surprised to find that k-fold CV is favored by the video-based studies over LOSO and LOUO. We expected the opposite due to the fact that video-based data is computationally expensive to work with and the aforementioned CV techniques require multiple runs of the same network. We assume k-fold CV is utilized to prevent bias in the dataset, hence, overfitting regardless of the expensive computational requirements.

Finally, we noted that the majority of the papers formulate skill assessment as a classification problem. There are only four papers that predict the performance score directly. However, the model developed by Kowalewski et al.[15] suggests $R^2$ of 0.03 with SD of 0.81; hence, we assume their model has limited or no generalization power. This leads us to believe that there is still room for improvement of regression accuracy. A reason for limited regression studies is, unlike classification, regression requires a ground-truth scoring system. However, such scoring systems are limited and not necessarily publicly available, e.g., the Fundamentals of Laparoscopic Surgery scores used for board certification are IP protected. In contrast, classification can be done based on their operative expertise.

We observed that intermediate surgeons are the most difficult to classify, which may have multiple reasons. First, the chosen categorization method may not reflect true surgical experience. For JIGSAWS, we hypothesize that there's a very large performance improvement that's seen between 10 and 100 hours of expertise, which might be making classification harder. Secondly, the DNN may be biased toward novice data. This is because expert behavior may not adequately represented due to an unbalanced size of the datasets, containing a significantly larger amount of novice data. This is particularly prevalent in the JIGSAW studies. Lastly, a surgeon may actually be an expert or a novice. Hence, any DNN classification for such cases would be considered correct by default.

Additionally, we observed the Intermediate surgeons were classified poorly, primarily for the Knot Tying task of JIGSAWS. This might be because the ground-truth performance of the experts and intermediates was reported to be similar to each other for the Knot Tying task in the literature.

## 5. Concluding remarks

In this paper, we examined papers related to the automation of surgical skill assessment using DNN. We used PRISMA guidelines to process papers and ended up analyzing 25 papers. As a result of the systematic review, we observed that DNN is useful to assess the surgical skills and has already reported higher accuracies comparing top ML models, with the work of Zia *et al.* being the only exception as tabulated in Table 3. Tool and hand motion data obtained from sensors and videos have been used for DNNs, though the use of other sensor-based metrics is possible. In recent work, we have shown that noninvasive brain imaging can be utilized to predict FLS scores in pattern cutting using a CNN framework named Brain-NET.[11] In our analysis CNN outperformed various machine learning techniques, namely, Kernel-Partial Least Squares, Support Vector Regression, and Random Forests.[11] This survey yielded that the best performing models all rely on convolutional neural networks, including the current state-of-the-art model in DL, achieving overall 98.7% accuracy on the JIGSAWS dataset.[39] Finally, we found that the most preferred method to report technical skills to classify surgeons into distinct categories, namely Expert, Intermediate, and Novice.

Several limitations of the current studies are evident from the analysis, which points to future research directions. The most important limitation is, of course, the lack of publicly available datasets of significant size. Surgical data is expensive to collect, and participation is voluntary. Having sufficient numbers of expert surgeons dedicate time to generate datasets is challenging due to their clinical obligations. Moreover, there is no consensus on the datasets regarding the effects of gender and handedness to the outcome. Thus, we expect to see future datasets to focus on the diversity as much as the quantity.

Moreover, we noticed a lack of rigor in establishing ground-truth performance scores. Most of the studies suggest they had an expert on the team who did the annotation, which might adversely affect inter-rater reliability. Moreover, some studies did not even cite the source of their supervision, meaning it is not clear whether they had consulted an expert or not to annotate the ground-truth performance. Additionally, the majority of the papers utilized kinematic data over video-based data, possibly due to limited video information in existing surgical datasets.

Moreover, there is only one end-to-end DNN model based on video analysis. With the advances in data augmentation to dilute the effect of sample size, we believe there will be more studies focusing on end-to-end frameworks via video sequences.

We were only able to find four regression studies in the literature. Regression differs from classification as the aim is to predict scores for each performance based on GRS. This makes regression advantageous as it is continuous, and the progress can be tracked by the trainee even when the assigned class for the performances does not change. Unlike regression, classification puts surgeons into discrete categories that leave gray areas. For example, for JIGSAWS, the self-proclaimed range of expertise was large for intermediates (10-100 hours), with novices up to 10 hours. Experts were 100+ hours, but we don't know the actual hours required to classify surgical skills. However, regression requires more data to pinpoint the performance. Also, there is not an agreement in the literature as to how should the different regression models be evaluated.[38]. We see this as a limitation for regressive models and expect more studies to be published in this field.

Finally, we found only one study based in the OR setting, while the rest of the studies either used physical or virtual simulators for skill assessment. While the training and certification of surgeons can be done on simulators, it is critical to be able to quantify their skills in clinical situations objectively. There are multiple challenges to doing this due to the inherent messiness of the operative environment where more than one surgeon may be involved in performing a procedure, annotation of the operative videos to provide high-quality training datasets and privacy issues related to the acquisition and use of the datasets.

In conclusion, we believe that DNNs possess strong potential for assessing surgical skills rapidly and objectively, providing real-time feedback, and certifying and credentialing surgeons in the near future. The availability of large, high-quality, publicly available annotated datasets will accelerate the achievement of that goal.

## 6. References

[1] Nataraja RM, Webb N, Lopez PJ. Simulation in paediatric urology and surgery, part 2: An overview of simulation modalities and their applications. Journal of pediatric urology. 2018 Apr 1;14(2):125-31.

[2] Dutta S, Krummel TM. Simulation: a new frontier in surgical education. Advances in surgery. 2006 Sep 1;40:249-63.

[3] Martin JA, Regehr G, Reznick R, Macrae H, Murnaghan J, Hutchison C, Brown M. Objective structured assessment of technical skill (OSATS) for surgical residents. British journal of surgery. 1997 Feb;84(2):273-8.

[4] Reiley CE, Lin HC, Yuh DD, Hager GD. Review of methods for objective surgical skill evaluation. Surgical endoscopy. 2011 Feb 1;25(2):356-66.

**Author biographies**

**Erim Yanik, M.S.,**
is a Ph..D. student in the Department of Mechanical, Aerospace and Nuclear Engineering. He received his Bachelor of Science degree in Mechanical Engineering from Istanbul Technical University and obtained his M.S. from Southern Illinois University Edwardsville. His research interests are Computer Vision, and Deep Neural Networks for surgical training and skill assessment.

**Xavier Intes, Ph.D.,**
is a professor in the Department of Biomedical Engineering, codirector of the Center for Modeling, Simulation, & Imaging in Medicine, and an AIMBE/SPIE/OSA fellow. He acted as the chief scientist of Advanced Research Technologies Inc., Montreal, Canada. His research interests are on the application of diffuse functional and molecular optical techniques for biomedical imaging in preclinical and clinical settings.

**Uwe Kruger, EngD.,**
is a Professor of Practice in Biomedical Engineering at RPI and engaged in research related to biomedical data science. Dr. Kruger is also a member of the Center for Modeling, Simulation and Imaging in Medicine where he provides oversight to projects involving statistical and machine learning problems. His research interests focus on developing and applying innovative solution methods to solve nonlinear regression and classification problems.

**Pingkun Yan, Ph.D.,**
is an Assistant Professor at the Department of Biomedical Engineering at Rensselaer Polytechnic Institute. He obtained his Ph.D. from National University of Singapore. His research interests include medical imaging informatics and image-guided intervention using machine learning and computer vision techniques.

**David Diller, Ph.D.,**
is a Senior Scientist at Raytheon BBN Technologies. He received his Ph.D. from Indiana University Bloomington. His research interests are AI, computer vision, and human performance modelling.

**Brian Van Voorst, M.S.,**

is a Lead Scientist at Raytheon BBN Technologies. He obtained his M.S. from Michigan Technological University. His research interests are AI and computer vision.

**Basiel Makled, M.S.,**

is a Science and Technology Manager and Engineer at the United States Army Combat Capabilities Development Command. His work is focused on Medical Simulation and Human Performance Training out of Orlando, FL. Basiel completed his Master of Science in Biomedical Engineering at the University of Central Florida and completed his undergraduate schooling at Florida State University where he received a Bachelor of Science in Engineering with Honors Distinction. Mr. Makled's primary research interests lie at the cross-roads of Medical Simulation, Training, Human Performance, Artificial Intelligence, and Virtual Reality.

**Jack Norfleet, Ph.D.,**

is a Chief Engineer at the United States Army Combat Capabilities Development Command. He received his degree from the University of Central Florida. His interests are medical simulations and human tissue properties.

**Suvranu De, Sc.D.,**

currently serves as the J. Erik Jonsson '22 Distinguished Professor of Engineering, Director of the Center for Modeling, Simulation and Imaging in Medicine and Head of the Department of Mechanical, Aerospace and Nuclear Engineering. His research interests are medical simulations, virtual reality, surgical training, machine learning